\documentclass{article}

\PassOptionsToPackage{numbers,square,sort,comma}{natbib}
\usepackage[preprint]{neurips_2021}

\usepackage{amsfonts}
\usepackage{amsmath}

\usepackage[utf8]{inputenc} 
\usepackage[T1]{fontenc}    
\usepackage{hyperref}       
\usepackage{url}            
\usepackage{booktabs}       
\usepackage{amsfonts}       
\usepackage{nicefrac}       
\usepackage{microtype}      
\usepackage{xcolor}         

\usepackage{graphicx}

\usepackage{multicol}
\usepackage{multirow}
\usepackage{booktabs}

\usepackage{listings,xcolor,courier,bookmark}
\usepackage{listingsutf8}
\definecolor{darkblue}{named}{blue}
\definecolor{darkred}{named}{red}
\definecolor{grau}{named}{gray}
\let\Righttorque\relax
\title{Few-shot Named Entity Recognition with Cloze Questions}

\author{
Valerio La Gatta, Vincenzo Moscato, Marco Postiglione, Giancarlo Sperlì \\
Department of Information Technology and Electrical Engineering, \\ University of Naples Federico II  \\
\{\texttt{valerio.lagatta, vincenzo.moscato,} \\ \texttt{marco.postiglione, giancarlo.sperli}\}\texttt{@unina.it}
}

\begin{document}

\maketitle

\begin{abstract}
  Despite the huge and continuous advances in computational linguistics, the lack of annotated data for Named Entity Recognition (NER) is still a challenging issue, especially in low-resource languages and when domain knowledge is required for high-quality annotations. Recent findings in NLP show the effectiveness of cloze-style questions in enabling language models to leverage the knowledge they acquired during the pre-training phase. In our work, we propose a simple and intuitive adaptation of Pattern-Exploiting Training (P{\small ET}), a recent approach which combines the cloze-questions mechanism and fine-tuning for few-shot learning: the key idea is to rephrase the NER task with patterns. Our approach achieves considerably better performance than standard fine-tuning and comparable or improved results with respect to other few-shot baselines without relying on manually annotated data or distant supervision on three benchmark datasets: NCBI-disease, BC2GM and a private Italian biomedical corpus. 
\end{abstract}

\section{Introduction}

    Recent advances in computational linguistics, characterized by an intensive study and wide adoption of language models \citep{Devlin2019BERTPO, Brown2020LanguageMA, Raffel2020ExploringTL}, have led to precious improvements in Named Entity Recognition (NER), which consists in identifying and classifying entities in a given text. However, the annotation of datasets for specific domains or languages is an expensive and time-consuming process, which often requires domain knowledge (e.g. healthcare, finance). At the present, few-shot settings for NER with pre-trained language models have not been extensively studied yet. 
    
    A recent line of research provides pre-trained language models with "task descriptions" in the form of \emph{cloze-questions} \citep{Radford2019LanguageMA}, which enable them to leverage the knowledge they acquired during the pre-training phase. Brown et al. \citep{Brown2020LanguageMA} show that this approach, without any sort of fine-tuning, allows to obtain high performance for a variety of SuperGLUE \citep{Wang2019SuperGLUEAS} tasks with only 32 examples. Schick et al. \citep{Schick2020ExploitingCQ} show that the extension of such approach with regular gradient optimization results in better and lighter models.
    
    To the best of our knowledge, the use of cloze-questions has not been investigated yet for NER tasks, probably due to their sequence-labeling nature: it is difficult (if not impossible) to provide a single task description which allows the language model to assign a label to each token in the input text. In this work, we propose a simple and intuitive adaptation which enables us to test the effectiveness of cloze-questions for NER: from the input example, we generate a new sentence for each token it contains. An example is shown below:
    \begin{quote}
        \emph{A diagnosis of SARS-CoV-2 was made in a young boy. In the sentence above, the word "young" is at the \lstinline{[MASK]} of a disease entity.}
    \end{quote}
    The language model will thus learn how to appropriately replace the masked token (e.g. \lstinline{[MASK]} = beginning/inside/outside).
    
    Being based on the Pattern-Exploiting Training (P{\small ET}) framework from Schick et al. \citep{Schick2020ExploitingCQ}, we will refer to our approach as P{\small ETER} (\textbf{P}attern-\textbf{E}xploiting \textbf{T}raining for Named \textbf{E}ntity \textbf{R}ecognition).

\section{P{\small ETER}: Pattern-Exploiting Training for Named Entity Recognition}
\begin{figure*}[t]
    \centering
    \includegraphics[width=\textwidth]{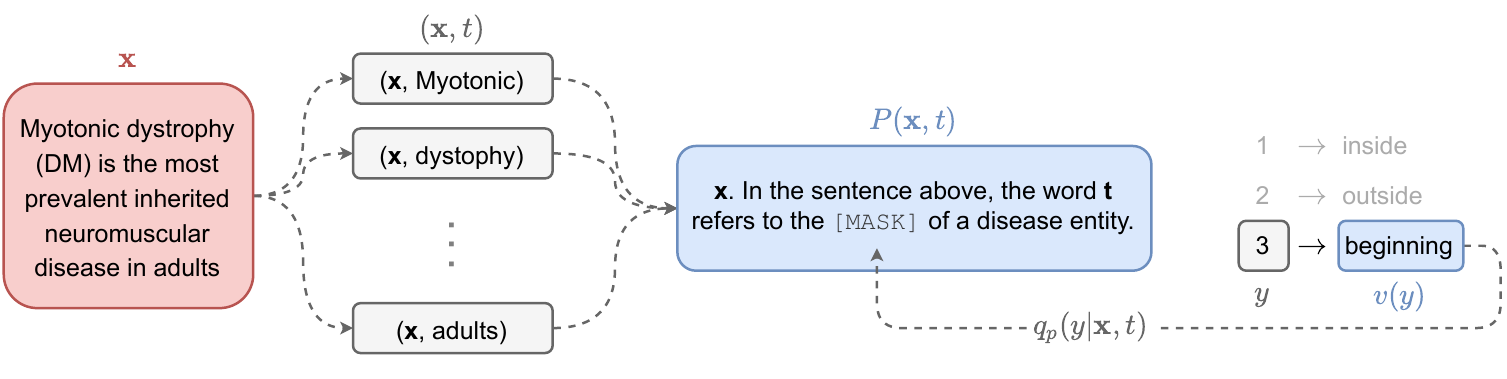}
    \caption{Example of PVP application. An input example $(\mathbf{x},t)$ is generated for each token $t$ in the sentence $\mathbf{x}$. A label $y$ is related to the token $t$. The pattern $P(\mathbf{x},t)$ and verbalizer $v(y)$ are then applied to generate the input of the masked language model, which is then fine-tuned on the $q_p(y|\mathbf{x}, t)$ loss.}
    \label{fig:PETER}
\end{figure*}

    We describe here the slight adjustment we made to enable P{\small ET} to handle the recognition of entities.

    Given a pre-trained masked language model $M$ with $T$ as its vocabulary, P{\small ET} addresses the task of mapping a textual input $x\in X$ to some label $y\in Y=\{1,...,k\}$ with $k\in \mathbb{N}$ by making $M$ predict the token which has to be replaced to a masked token \lstinline{[MASK]} $\in T$. To accomplish this task, P{\small ET} requires a set of \emph{pattern-verbalizer pairs} (PVPs) consisting in:
    
    \begin{itemize}
        \item a \emph{pattern} $P: X \rightarrow T^*$ which converts inputs to cloze questions, i.e. sequences of tokens containing exactly one \lstinline{[MASK]} token;
        \item a \emph{verbalizer} $v: Y \rightarrow T$ mapping labels to tokens representing their meanings.
    \end{itemize}
    
    The training set $\mathcal{T}$ for supervised NER consists in a list of couples $\{(\mathbf{x_1}, \mathbf{y_1})\}_{i=1}^{\left|\mathcal{T}\right|}$, where the $i$-th couple is a sequence of token-tag pairs $(x,y)$. The set of tags depends on the NER schema (e.g. IO, IOB2, IOBES). Differently from P{\small ET}, our patterns take in input not only the sentence but also the token which the cloze-question refers to, as shown in Figure~\ref{fig:PETER}. The steps of our pipeline are listed below:
    
    \begin{enumerate}
        \item from the textual input $\mathbf{x}$, $\left|\mathbf{x}\right|$ input examples are generated, each of them referring to a different token in the sentence;
        \item for each PVP $\langle	P(\mathbf{x},t),v(\mathbf{y})\rangle$, a masked language model $M$ is fine-tuned based on the conditional probability distribution of $y$ given $(\mathbf{x},t)$:
        \begin{equation}
            q_P(y|\mathbf{x}, t) = \frac{\text{exp} M(v(y) | P(\mathbf{x},t))}{\sum_{i=1}^k\text{exp}M(v(i) | P(\mathbf{x},t))},
        \end{equation}
        were $M(v(y) | P(\mathbf{x},t))$ denotes the score that the language model assigns to the token at the masked position of being $v(y)$;
        \item unlabeled data are used to generate a soft-labeled dataset from the aggregation of predictions of the previously trained masked language models;
        \item the soft-labeled dataset is used to train a masked language model with a classification head.
    \end{enumerate}

\section{Experiments}

  
        \noindent \textbf{Datasets} 
            We used the original train-dev-test splits to facilitate comparability. However, in accordance with Schick et al. \citep{Schick2020ItsNJ}, we do not use development sets and we do not perform hyperparameter optimization, due to its non-feasibility in few-shot scenarios. For three replications with different seeds, we randomly select $k$ training instances from the training split ($k \in \{10,25,50,100\}$ being the number of "shots") and report results on the whole test set. 
            We evaluate our framework based on the following datasets:
            
            \begin{itemize}
                \item \emph{NCBI-Disease} \citep{Dogan2014NCBIDC}: collection of 793 fully annotated PubMed abstracts with 790 unique disease concepts. Test set contains 960 entities, while 10-, 25-, 50- and 100-shot training sets contain and average of 8, 27, 49 and 95 entity mentions, respectively.
                \item \emph{BC2GM} \citep{Smith2008OverviewOB}: Test set contains 6325 entities, while 10-, 25-, 50- and 100-shot training sets contain and average of 14, 32, 58 and 109 entity mentions, respectively.
                \item \emph{WINCARE}: We used a private Italian dataset of cardiological clinical notes which has been provided by \emph{Azienda Ospedaliera Universitaria Federico II}. Test set contains 845 disease entities, while 10-, 25-, 50- and 100-shot training sets contain and average of 10, 27, 52 and 106 entity mentions, respectively.
            \end{itemize}
            
            We use the pre-processed versions of NCBI-Disease and BC2GM provided by Wang et al. \citep{Wang2019CrosstypeBN}.
        
        \noindent \textbf{Patterns}
            We experiment with the following patterns:
            \begin{itemize}
                \item $P_1(\mathbf{x},t)$: $\mathbf{x}$. In the sentence above, the word $t$ refers to the \lstinline{[MASK]} of a \emph{<entity type>} entity.
                \item $P_2(\mathbf{x},t)$: $\mathbf{x}$. Question: In the passage above, which part of a \emph{<entity type>} entity does the word $t$ refers to? Answer: \lstinline{[MASK]}.
            \end{itemize}
            
            Note that \emph{<entity type>} is replaced with "disease" in the first and second datasets and "gene" in the second one.
            
        \noindent \textbf{P{\small ET}} 
            We have mostly left the default configurations of PET unchanged, with the exception of the following settings: (1) we set the maximum sequence length to 128 for NCBI-Disease and BC2GM and to 256 for ClinicalNotesITA; (2) we use BioBERT \citep{Lee2020BioBERTAP} for NCBI-Disease and BC2GM and GilBERTo\footnote{https://github.com/idb-ita/GilBERTo} for ClinicalNotesITA as pre-trained language models; (3) we limit the number of unlabeled examples to 10.000 for a faster training; (4) we set the number of epochs to 10, 7, 5, 5 for the 10-, 25-, 50- and 100-shot scenarios, respectively.
            
        \noindent \textbf{Metrics} 
            We use \emph{precision}, \emph{recall} and \emph{f1} scores to evaluate our results. We refer to \lstinline!seqeval!\footnote{\url{https://github.com/chakki-works/seqeval}} and \lstinline{sklearn}\footnote{\url{https://scikit-learn.org}} for evaluations on the IOB2 schema.

        \begin{figure}
            \centering
            \includegraphics[width=\textwidth]{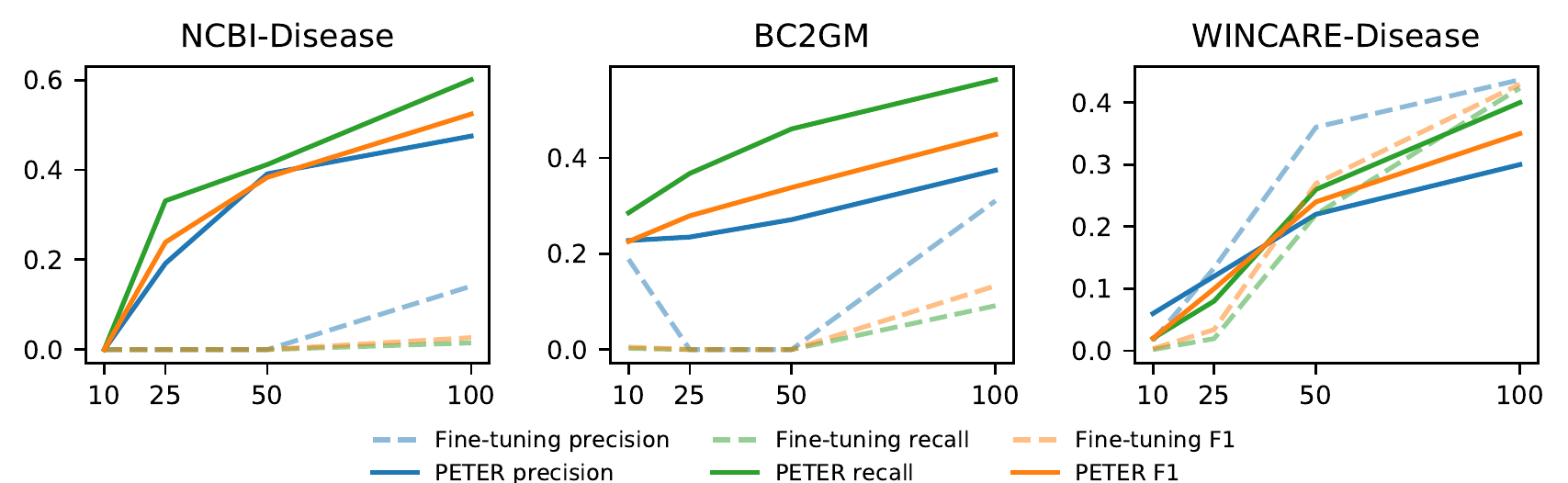}
            \caption{Average precision, recall and F1 scores obtained with the standard fine-tuning approach and PETERas the number of shots increase}
            \label{fig:vs_finetuning}
        \end{figure}

        \begin{table}
  \caption{Mean and standard deviation of precision (P), recall (R) and F1 scores in the 10-shot scenario.}
  \label{tab:vs-triggerNER}
  \centering
  \begin{tabular}{lllll}
    \toprule
    Dataset & Method & Precision & Recall & F1 \\
    \midrule
    \multirow{4}{*}{NCBI-Disease}
     & F{\small INE}T{\small UNING} & $0$ & $0$ & $0$ \\
     & T{\small RIGGER}NER  & $0.01\pm 0.02$ & $0.01\pm 0.02$ & $0.01\pm 0.02$ \\
     & BOND & $\mathbf{0.02\pm 0.03}$ & $0.01\pm 0.01$ & $0.01\pm 0.01$ \\
     & P{\small ETER} (ours) & 0 & 0 & 0   \\ \midrule
    \multirow{4}{*}{BC2GM}
     & F{\small INE}T{\small UNING} & $ 0.1882 \pm 0.1930 $ & $ 0.0026 \pm 0.0036 $ & $ 0.0060 \pm 0.0083 $ \\
     & T{\small RIGGER}NER  & $0.10\pm 0.06$ & $0.01\pm 0.01$ & $0.02\pm 0.02$  \\
     & BOND & $0.10\pm 0.05$ & 0 & $0\pm 0.01$ \\
     & P{\small ETER} (ours) & $\mathbf{0.24\pm 0.27}$ & $\mathbf{0.22\pm 0.26}$ & $\mathbf{0.23\pm 0.27}$   \\ \midrule
    \multirow{4}{*}{WINCARE}
     & F{\small INE}T{\small UNING} & $ 0.0162 \pm 0.0129 $ & $ 0.0019 \pm 0.0014 $ & $ 0.0032 \pm 0.0024 $ \\
     & T{\small RIGGER}NER  & $0.01 \pm 0.01$ & $\mathbf{0.05\pm 0.04}$ & $0.02\pm 0.01$  \\
     & BOND & 0 & 0 & 0 \\
     & P{\small ETER} (ours) & $\mathbf{0.06\pm 0.04}$ & $0.02\pm 0.03$ & $0.02\pm 0.03$  \\
    \bottomrule
  \end{tabular}
\end{table}

            \noindent \textbf{Results} To gauge the ability of P{\small ETER} to provide effective annotations with few examples, we compared its results with (1) the regular fine-tuning approach, consisting in starting from a pre-trained language model and fine-tuning with a classification head, (2) TriggerNER \citep{Lin2020TriggerNERLW}, which uses manually annotated triggers to guarantee good few-shot performance and (3) BOND \citep{DBLP:journals/corr/abs-2006-15509}, which leverages distant supervision to increase automatically the labelled samples. 
            
            Figure~\ref{fig:vs_finetuning} shows a significant improvement with respect to the standard fine-tuning approach for NCBI and BC2GM datasets. In the case of the WINCARE dataset, P{\small ETER} has similar performance to the baseline, probably due to the fact that all the WINCARE clinical notes are collected from the same department of cardiology, and thus could be effectively represented by few instances. 
            
            In Table~\ref{tab:vs-triggerNER} we report results of baseline models in a highly-constrictive 10-shot scenario. P{\small ETER} results are extremely promising and worthy for further research: our approach can indeed reach comparable or superior performance with respect to other baselines without relying on manual annotated triggers, as in T{\small RIGGER}NER, or distant supervision, as in BOND.

\section{Related Work}

    Recently, there is an increasing number of research works about few-shot Named Entity Recognition. They are mainly based on \emph{meta-learning} \citep{Li2020MetaNERNE, Krone2020LearningTC, Oguz2021FewshotLF, Fritzler2019FewshotCI, Yin2020KnowledgeawareFL, Cong2020FewShotED}, \emph{distant} and \emph{weak supervision} \citep{Adelani2020DistantSA, Lan2020LearningTC, Liu2019KnowledgeAugmentedLM, Lou2020AGA, Shang2018LearningNE, Zeng2020CounterfactualGA} and \emph{transfer learning} \citep{Bondarenko2020UsingFL, Reimers2020MakingMS, Schneider2020BioBERTptA}. However, current methods usually rely on additional data or annotating efforts \citep{Lin2020TriggerNERLW}, which can be a strong limitation in low-resource languages and domains.

    A recent line of research uses cloze-style questions to rephrase tasks, enabling language models to obtain high performance in unsupervised scenarios by leveraging the knowledge acquired during the pre-training phase \citep{Radford2019LanguageMA}. This idea has been successfully applied to unsupervised text classification \citep{Puri2019ZeroshotTC}, commonsense knowledge mining \citep{Feldman2019CommonsenseKM} and knowledge probing \citep{Petroni2019LanguageMA, Ettinger2019WhatBI}. Schick et al. \citep{Schick2020ExploitingCQ} show that not only does the cloze-question approach can be combined with regular gradient-based fine-tuning, but it can also lead to considerable performance improvements, in both quality and efficiency terms. Thanks to the inclusion of few supervised data to fine-tune the language model, it is possible to obtain comparable performance with GPT-3 \citep{Brown2020LanguageMA} while using far fewer parameters \citep{Schick2020ItsNJ}.
    
    Being originally designed for text classification, to the best of our knowledge, the use of cloze-questions for NER in few-shot settings has not been investigated until now.
    
\section{Conclusion}
    We have shown that the cloze-questions mechanism can be also a valuable technique for few shot NER. Our approach, based on an adaptation of P{\small ET}, strongly out-performs standard fine-tuning and obtains comparable results with the state-of-the-art without relying on any additional information, such as additional manual annotations or distant supervision. In future work, we would like to extend this approach for multi-entity datasets and to further investigate the application of cloze-questions for NER.

\bibliography{anthology}
\bibliographystyle{plain}

\end{document}